\def\year{2022}\relax
\documentclass[letterpaper]{article} 
\usepackage{aaai22}  
\usepackage{times}  
\usepackage{helvet}  
\usepackage{courier}  
\usepackage[hyphens]{url}  
\usepackage{graphicx} 
\usepackage[sort]{natbib}  
\usepackage{caption} 
\DeclareCaptionStyle{ruled}{labelfont=normalfont,labelsep=colon,strut=off} 
\newcommand{\citepos}[1]{\citeauthor{#1}'s (\citeyear{#1})}
\usepackage{booktabs}
\usepackage[inline]{enumitem}
\newlist{myenumerate}{enumerate*}{1}
\setlist[myenumerate]{label=\emph{(\arabic*)}, after={.}, itemjoin={{; }}, itemjoin*={{; and }}}

\title{Consent as a Foundation for Responsible Autonomy}

\author{Munindar P.~Singh}

\affiliations{
  Department of Computer Science\\
  North Carolina State University\\
  Raleigh, NC 27695-8206, USA\\
  singh@ncsu.edu
}

\begin{document}

\maketitle

\begin{abstract}
This paper focuses on a dynamic aspect of responsible autonomy,
namely, to make intelligent agents be responsible at run time.  That
is, it considers settings where decision making by agents impinges
upon the outcomes perceived by other agents.  For an agent to act
responsibly, it must accommodate the desires and other attitudes of
its users and, through other agents, of their users.

The contribution of this paper is twofold.  First, it provides a
conceptual analysis of consent, its benefits and misuses, and how
understanding consent can help achieve responsible autonomy.  Second,
it outlines challenges for AI (in particular, for agents and
multiagent systems) that merit investigation to form as a basis for
modeling consent in multiagent systems and applying consent to achieve
responsible autonomy.

\end{abstract}

\section{Introduction}
\label{sec:introduction}

Recent advances in the capabilities of AI have rightly brought the
concerns of responsible AI to the forefront.  Mainstream AI efforts
consider topics such as algorithmic accountability
\citep{Diakopoulos-16:accountability} and fairness, referring to
statistical properties of AI algorithms \citep{Leben-20:fairness-ML}.
These approaches consider the application of AI in different settings,
such as assistance in judicial sentencing guidelines or loan
application processing, and bring out the broader societal context.
However, in those cases the context is largely fixed---we would not be
able to easily change the judicial or financial systems, and get rid
of the disparities and inequities entrenched in those systems.

Research into responsibility in AI focuses on static aspects such as
design methodologies and development practices for AI
\citep{Dignum-17:responsible,Dignum-19:responsible}, e.g., via
checklists for agent developers.  In contrast, although such
design-time aspects of responsible AI are undoubtedly valuable, we
focus here on \emph{responsible autonomy}, which we define as
including the challenges of ensuring that an autonomous agent
exercises its autonomy responsibly, i.e., at runtime.

There are increasing prospects of people contesting decisions made by
(or using) AI, especially when an AI agent's decisions affect people
without being mediated by a human.  Consent would be a crucial element
of any justification of agent decisions on ethical and legal grounds

\subsection{Sociotechnical Systems}

In our formulation of responsible autonomy, we consider agents
embedded in microsocieties, wherein the context is modeled
computationally and can be reasoned about and potentially controlled
by the agents.  We model such a microsociety as a \emph{sociotechnical
  system (STS)} comprising autonomous social entities (people and
organizations or \emph{principals}) and technical entities (software
or \emph{agents}, who help principals)
\citep{TOSEM-20:Desen,TIST-13-Governance}.  Simply put, we place
agents in a social setting.  Designers of STSs and agents and the
agents themselves at runtime represent and reason about the social
setting provided by an STS.  A classical means to model the relevant
parts of the social setting is through deontic constructs called
\emph{norms}.  Norms in this sense encompass both social norms of the
ordinary vernacular and legal constructs
\citep{Von-Wright-63:Norm,Von-Wright-99:Personal}.

We posit that understanding responsible autonomy from a dynamic
perspective will facilitate responsible AI in static settings by
making explicit what is otherwise hidden.

\subsection{Contributions}
We formulate consent from an STS standpoint to address responsible
autonomy.  We identify crucial criteria for consent based on a brief
review of the relevant literature.  We articulate a way of
understanding consent in AI that accommodates both the social and the
technical architecture of an STS.  Our overarching claim is that
consent provides a new foundation for approaching responsibility that
highlights autonomy, social interaction, and accountability.

Incorporating consent in AI is not only practically important as AI
becomes more capable (and open to contestation), but also exposes
important research questions.  Specifically, consent lies at the heart
of autonomy \citep{Alexander-96:consent,Hurd-96:consent} and our STS
formulation ties it to responsible autonomy in individual and social
action.  The ensuing research questions and require participation by
AI research communities in multiagent systems, human-AI systems, agent
communication, and agent architectures with additional opportunities
in dialogue understanding and active learning.

\subsubsection{Organization}

Section~\ref{sec:Responsible} introduces governance.
Section~\ref{sec:Uses} describes how consent in used in computing,
law, and philosophy.  Section~\ref{sec:Challenges} summarizes key
challenges for consent.  Section~\ref{sec:Vision} discusses our vision
for future research.

\section{Responsible Autonomy and Governance}
\label{sec:Responsible}

We contrast consent-based governance for autonomy.

\subsection{Moral Quandaries}
Besides statistical properties of algorithms, a second dominant theme
in research into AI ethics concerns agent decision making, focusing on
moral quandaries an agent may face.  These quandaries are formulated
in contexts where an agent's decision has outcomes (on third parties)
with ethical import.  The canonical such quandary is where a (train)
trolley is running on some tracks and the agent has the option to
switch it to an alternative pair of tracks \citep{Foot-67:trolley}.
People of differing numbers and attributes are tied to each pair of
tracks.  Thus, the agent's decision would lead to saving some and
killing some.  The term ``trolley problem'' is often used generically
for all such moral dilemmas, many of which do not involve vehicles
\citep{Wood-11:Parfit}.

\citet{Fried-12:trolley} characterizes trolley problems as posing
hypothetical dilemmas where (1) the agent is an individual, not an
institution; (2) the agent faces a one-off decision; (3) the causal
chain between decision and outcome is short; (4) the consequences of
the decision are known \emph{a priori} with certainty; and (5) there
are no opportunities for scaling up the moral reasoning or its
principles to a larger number and variety of cases.  In essence, the
trolley problems are rigged to be context-free situations seeking to
highlight and promote a non-consequential viewpoint, downplaying any
aggregative (utilitarian) approach to reasoning about ethics and risk.

But realistic AI approaches, just like human decision making
\citep[p.~506]{Fried-12:trolley}, must accommodate uncertainty and
trade off risks in the long run.  Thus, we posit that focusing on
moral quandaries has little to offer in the way of valuable research
questions for responsible autonomy.

\subsection{Governance}
Responsible autonomy means not only promoting one's own values and
preferences but also refraining from violating the values and
preferences of others.  In addition to producing agents who are
individually responsible, we want to provide system-level guidance for
responsibility.  This is the province of \emph{governance} in STS
\citep{Baldoni+21:accountability,TIST-13-Governance}, not to be
confused with offline designs or administrative processes through
which agents may be coordinated.  That is, governance concerns how
agents operate and interact to achieve system-level objectives while
satisfying their users' needs.  A motivation for this formulation of
governance is that in complex contexts, it is not viable to produce
constraints that are inflexible and effective; therefore, the agents
much govern themselves (i.e., each other).  Such flexibility is
especially valuable when an agent's decisions affect the outcomes for
others, that is, when responsibility matters.

In an STS setting, the challenge of achieving responsible autonomy
splits into two parts, corresponding respectively to macro and micro
ethics \citep{AIES-18:ethics}:
\begin{itemize}
\item Specifying an STS based on the requirements that minimizes
  unethical outcomes as judged by its stakeholders.
\item Specifying agents who behave in a way compatible with the values
  of their stakeholders and the norms of the STS.
\end{itemize}

\subsection{Consent-Based Governance}
Both of the above components of responsible autonomy rely upon an
approach for governance based on a proper understanding of
\emph{consent}.  Specifically, our motivation for placing consent at
the center of responsible autonomy is that consent is a foundational
construct in autonomy, both in terms of exercising one's autonomy and
in recognizing the autonomy of others, the latter being an essential
element of Kantian ethics \citep[p.~90]{Hill-80:Kant-humanity}.

As an illustration of responsible autonomy via consent, suppose an
agent assists its user, Alice, in taking actions that promote Alice's
goals.  Consider a mobile social application that supports sharing
one's picture or location.  The problem is social because one user's
action affects another's privacy
\citep{Kurtan+Yolum-21:privacy,Mosca+20:privacy}, such as when a joint
picture is shared.
When Alice's agent acts on behalf of her, it affects outcomes for her
and for other people.  For example, when posting a picture of Alice
with her friends, the agent must act responsibly regarding the
wellbeing both of Alice and of the friends.  The agent may have a
formal fiduciary duty toward Alice and at least an informal fiduciary
duty toward her friends, the latter reflecting her moral duty not to
exploit her friends.

Consent is a natural abstraction here.  In informal terms, the agent
should ensure that the people whose picture is being shared consent to
the sharing.  The agent may have previously obtained consent from
Alice and must explicitly or otherwise obtain consent from the others.
More generally, for ethical behavior, ideally, any party affected
adversely, or potentially so, must consent to the first party's
actions.  How can we specify the sharing microsociety (possibly
realized via the app) so that the values of the users are respected?
How can we build such an agent to apply the norms of the corresponding
STS?

Surprisingly, however, consent has not been studied in AI.  That is,
consent is applied in a purely uninterpreted manner without any way to
represent and reason about it.

\section{The Many Uses of Consent}
\label{sec:Uses}

We review some key applications of consent.

\subsection{Moral and Legal Legitimacy}
Understanding consent is crucial in assessing the legitimacy of an
action, which would determine whether a crime took place.  For
example, consent is the difference between borrowing and stealing and
between lovemaking and rape.  The volenti maxim is that an explicit
consent (or request implying consent) overrides ordinary prohibitions
\cite{Dempsey-18:consent}.  Consent characterizes when some action by
one autonomous party gains legitimacy despite potentially infringing
upon the autonomy or authority of another party---that is its ``moral
magic'' \citep{Alexander-96:consent,Hurd-96:consent}.

\subsection{Consent of the Governed}
Political philosophy has one of the oldest traditions in consent, going
back to John Locke, and using the ``consent of the governed'' as the
basis for the legitimacy of government.  This topic is relevant to AI
because the membership of an agent in an STS subjects the agent to the
norms of that STS (relative to its role in the STS) and, therefore,
the agent must consent to play a specific role in that STS.

\citet{Pitkin-65:consent-I,Pitkin-66:consent-II} describes how consent
of the governed relates to their political obligation to obey the
government.  She brings forth challenges in how the traditional
(Lockean) notion of consent may be applied.  Specifically, Pitkin
makes a case that the concept proves to be vacuous in human societies.
On the one hand, Locke argues consent by an individual is essential
for that individual to be subject to the laws of society.  On the
other hand, he postulates tacit consent derived merely from living in
a jurisdiction: that is, nonconsenting individuals (competent adults,
for the sake of simplicity) are governed the same as consenting
individuals.

Pitkin advocates for the alignment of values as the crucial point in
that an agent ought not to merely think of their consent as conferring
legitimacy but of an evaluation of the moral nature of the government
as conferring legitimacy.  In this regard, she also calls out an
emphasis on the opposite of obeying under consent, namely, a duty to
resist tyranny.

\subsection{Business Contracts}

Consent turns up as a basis for modeling business contracts to capture
the idea that the contracting parties enter into contracts freely and
to avoid some of the challenges arising in accounts based on (1)
intent or expectation; (2) efficiency or fairness; and (3) enforcement
processes \citep{Barnett-96:consent-contract}.  Consent provides a way
to model the social context and capture what each party's entitlements
are in that context and how they are assigned through a contract.  An
illustration would be in arbitration clauses through which the
contracting parties agree to waive a jury trial in case of a dispute.

Consent accords well with \emph{relational contracts}
\citep{Bernstein-93:social-norms,Bernstein-15:relational}, which focus
on social relationships between contracting parties.  By modeling the
social context, relational contracts accommodate renegotiation on the
fly to avoid disputes that invariably arise because no contract can
specify all possible eventualities.

\subsection{Consent in Computing}
\label{sec:consent-computing}

Established practice in computing regarding consent goes back to work
on privacy early in the information age.  \citet{Westin-67:privacy}
established the influential doctrine of \emph{notice and choice},
under which all you need to do to respect an individual's privacy
rights is to (1) disclose what information of theirs you are
obtaining, storing, using, or sharing, and (2) ask them to consent to
that action.

Notice and choice spread because it is easy to implement.  One usage
is in click-wrap licenses for a software product where a user is
provided a license and must accept its terms before accessing the
product.  Likewise, many websites demand that users consent to being
tracked.  And, social media apps have users consent to their
information being analyzed and shared with third parties.

But users are ill-equipped to figure out the ways their information
may be shared, analyzed, and combined with other information.
Therefore, notice and choice has been criticized by privacy scholars
\citep{Schermer+14:consent,Sloan+Warner-14:notice}.
\citet{Nissenbaum-04:integrity} makes a case for understanding the
contexts in which information is shared and developing privacy norms
accordingly.  Notice and choice hides the contexts and blind sides the
user, leading to their information propagating across contexts that
are decoupled in their mind.

Notice and choice naively assumes that people are rational
\citep{Hoofnagle+Urban-14:Westin}.  But people consent to whatever
terms are offered because of complexity and the power differential
(the terms being non-negotiable) \citep{Fassl+21:consent-theater}.
Recent work in human-computer interaction seeks improved models of and
user interfaces for consent
\citep{Im+21:affirmative-consent,Lindegren+21:consent-HCI}, but this
work doesn't offer much for responsible autonomy.

All considered, consent in computing is often misguided or
ill-intentioned, and does not shed light on responsibility.

\section{Challenges to Understanding Consent}
\label{sec:Challenges}

Intuitions about consent are far from established
\citep{Schnueriger-18:consent}.  First, consent reflects a mental
action of the consenting party---indicating that it is the exercise of
an internal choice.  Second, consent reflects a communicative act or
performative by the consenting party conferring powers on the
recipient---indicating that it is the exercise of a normative power
\citep{Hohfeld-19,Koch-18:consent,Hurd-18:consent}.  The mental
approach doesn't explain how a normative power arises from an internal
action without a communication.  The communicative approach doesn't
explain the treatment of mistakes in performing a communication that
grants consent.  \citet{Alexander-14:consent} distinguishes
\emph{wrongdoing} (causing harm by acting without there being true
mental consent on part of the party whose consent was necessary) from
\emph{culpability} (acting without belief that the requisite consent
exists).  These distinctions are important for ascribing blame.
Table~\ref{tab:consent} summarizes important criteria, using a
grouping explained next.

\begin{table}[htb]
\centering
\begin{tabular}{l p{2.434in}}\toprule
Criterion & Example or Explanation \\\midrule

Visibility & Consent is observable, i.e., a communication\\
Free will & Consenter acts without being coerced\\
Truth & Consenter's beliefs are true and complete \\\midrule
Capacity & Consenter is mentally fit \\
Cognition & Consenter believes and intends to\\
Attention & Consenter exercises mental faculties \\\midrule
Statutes & Consenter meets statutory criteria, e.g., age\\
Power & Consenter is not subjugated by consentee\\
Honesty & Consentee does not mislead consenter\\\bottomrule
\end{tabular}
\caption{Important criteria in consent grouped as
  \citepos{Habermas-84} objective, subjective, and practical validity
  claims.}
\label{tab:consent}
\end{table}

\section{Vision: Research on Consent in AI}
\label{sec:Vision}

Understanding consent is not only a prerequisite for achieving
responsible autonomy, it is also a subtle concept that melds
intuitions regarding ethics, law, usability, and decision making.  Our
sociotechnical stance reveals important opportunities for AI research.
We first outline ideas for a semantics of consent and then some
promising research directions.

\subsection{Toward a Social Semantics of Consent}
Traditional disputes, as between the objective and subjective elements
of consent, arise because of a confusion of meanings and meaning
standards.  We propose to apply \citepos{Habermas-84} framework to
reconcile these ideas by building on a public semantics for agent
communication \citep{Ijcai-99-ACL}, which was an adaptation of
Habermas.  Specifically, we enhance the original nonmentalist
conception of communication
\citep{Austin62,Sbisa-07,Sbisa-18:speech-norms}.  We develop validity
criteria for consent from the perspectives of the consenter, the
consentee, and a third party (potentially the STS in which consenter
and consentee interact).

\citepos{Habermas-84} theory of communication in the public sphere
associates three \emph{validity claims} (i.e., distinct standards of
meaning) with each communication: \emph{objective} (true);
\emph{subjective} (appropriate beliefs and intentions);
\emph{practical} (justified in the social context).  We relate these
claims to the criteria in Table~\ref{tab:consent}.  Objectively,
granting consent is a social action and its meaning is for the
consenter to empower the consentee, e.g., by forgoing any moral
objections to the consentee acting as specified.  It is valid under
visibility and free will.  Subjectively, consent is a mental object
and granting consent an intentional action.  Its meaning is the
corresponding intention.  It is valid provided it is performed
with a capacity to reason about consent, holds the right cognitive state
under full attention, and the consenter's beliefs pertinent to the
consent are true and include relevant facts.  Practically, granting
consent is valid if the consent is not prohibited by statutes (i.e.,
norms of the STS), the consenter is not subjugated by the consentee or
misled by the consentee.

\subsection{Bridging AI Ethics and Law}
Consent is key in distinguishing right from wrong and in legitimizing
actions and making them legal.  Thus, it opens research challenges on
bridging the gap between ethics and law to develop responsible agents.

\emph{Legal positivism} is the doctrine that the law is as it is posited.
Variants of positivism take stronger or weaker stances on these key
theses
\citep{Himma-21:legal-positivism,Green+Adams-19:SEP-legal-positivism}:
\begin{myenumerate}
\item \emph{pedigree}, that the law's existence and validity rely upon
  social facts, i.e., that it is declared a law
\item \emph{separability}, that though the law and morality may align,
  it is not necessary that they do
\item \emph{fallibility}, that the law may be intended to be moral and
  yet be deficient in that regard
\item \emph{neutrality}, that even though the law is not value
  neutral, it should be described and argued about in value-free terms
\item \emph{discretion}, that judges may exercise discretion where the
  law is not clear and in doing so, they extend the law
\end{myenumerate}

Natural law is an older doctrine that the law derives its legitimacy
from being \emph{natural}, i.e., granted by nature or by divine right.
In modern versions, it is the view that ethics be incorporated into
the law \citep{Gavison-82:jurisprudence}.  Modern judicial practice
focuses on ``applying the law'' and avoiding justifications based on
ethics.  That is, natural law has largely been supplanted by legal
positivism, whose main theses summarized above sit largely in contrast
with natural law.

Consent today is conceived of legally positively, i.e., based on a
consenter's utterances.  But its shortcomings arise mainly because
this thinking disregards ethics: is consent right if obtained from
someone who is desperate?  The discretion thesis of legal positivism
with practical validity from Habermas provides a potential opening in
how we might formulate consent so that it bridges the gap with ethics.

\subsection{Verifying Agents and STSs for Consent}

Responsible autonomy presupposes that an STS would provide social and
technical controls to promote responsible actions by its members and
to limit harm in case of malfunction or malfeasance.  How can we
verify an STS and its member agents to ensure that they provide
consent where needed, refrain from doing so where it is not needed,
and respect the absence of consent from others in their decisions?
For an STS, in addition, we need ways to (1) minimize the risk of an
agent being placed in an ethical quandary when providing or receiving
consent; and (2) ensuring that social and technical controls on agent
behavior balance any propensity to violate another's consent.

The literature on consent takes a \emph{retrospective} view, as to
adjudicate on a violation in a court of law.  For AI ethics, the
\emph{prospective} view of consent and responsibility
\citep{Van-de-Poel-11:responsibility} is no less important, to assess
an agent's decisions about when to consent and when to obtain and act
based on another's consent.  This direction would lead to formal
reasoning for verifying and certifying
\citep{Dennis+16:ethical,Fisher+21:certification} agents and STSs with
respect to consent.

\subsection{Consent-Based Design}
This paper expands the ontology of requirements beyond goals and
dependencies \citep{Yu+11}, legal norms such as social commitments
\citep{RE-14:Protos} and powers \citep{TIST-13-Governance}, and values
\citep{Cranefield+17:values-BDI} to give first-class status to
consent.  How would we produce suitably equipped design methodologies
that use consent to express requirements?  Such a methodology would
accommodate the criteria of Table~\ref{tab:consent}.  For example,
using these criteria, a suitable methodology would help model
\emph{consentability}---the power accorded within an STS to its
participants regarding the kinds of consent they can issue
\citep{Kim-19:consentability} that has the appropriate pragmatic
consequences in enabling further action by other participants or their
agents.

\subsection{Learning and Interaction about Consent}
An important direction is to model agent-user dialogue so that an
agent can elicit its user's consent, obtain valid consent (as
described above), and explain its decisions in light of consent from
its user as well as from other agents (on behalf of their users).  In
this light, how might we extend research on values
\citep{Liscio+21:Axies} to develop methods by which an agent
understands its user's consent?  To evade the criticisms of
traditional methods as discussed in
Section~\ref{sec:consent-computing}, a desirable approach must be
explicit yet not tedious, even when the consent is nuanced and
contextual.

\section*{Acknowledgments}
Thanks to the NSF (grant IIS-2116751) and to the DoD (Science of
Security Lablet) for support.

\DeclareRobustCommand{\nUmErAL}[1]{#1}\DeclareRobustCommand{\nAmE}[3]{#3}

\end{document}